\documentclass[conference]{IEEEtran}
\usepackage{times}
\usepackage{xcolor}

\usepackage{graphicx}

\usepackage[numbers]{natbib}
\usepackage{multicol}
\usepackage[bookmarks=true]{hyperref}
\usepackage{amsmath} 
\usepackage{amssymb}
\usepackage{multirow}
\usepackage{caption}
\usepackage{subcaption}
\usepackage[normalem]{ulem} 

\usepackage{layout}

\begin{document}

\title{MULAN-WC: Multi-Robot Localization Uncertainty-aware Active NeRF with Wireless Coordination}

\author{Author Names Omitted for Anonymous Review. Paper-ID [add your ID here]}
\let\oldtwocolumn\twocolumn
\renewcommand\twocolumn[1][]{%
    \oldtwocolumn[{#1}{
    \begin{center}
           \includegraphics[width=0.75\textwidth,height=7.5cm]{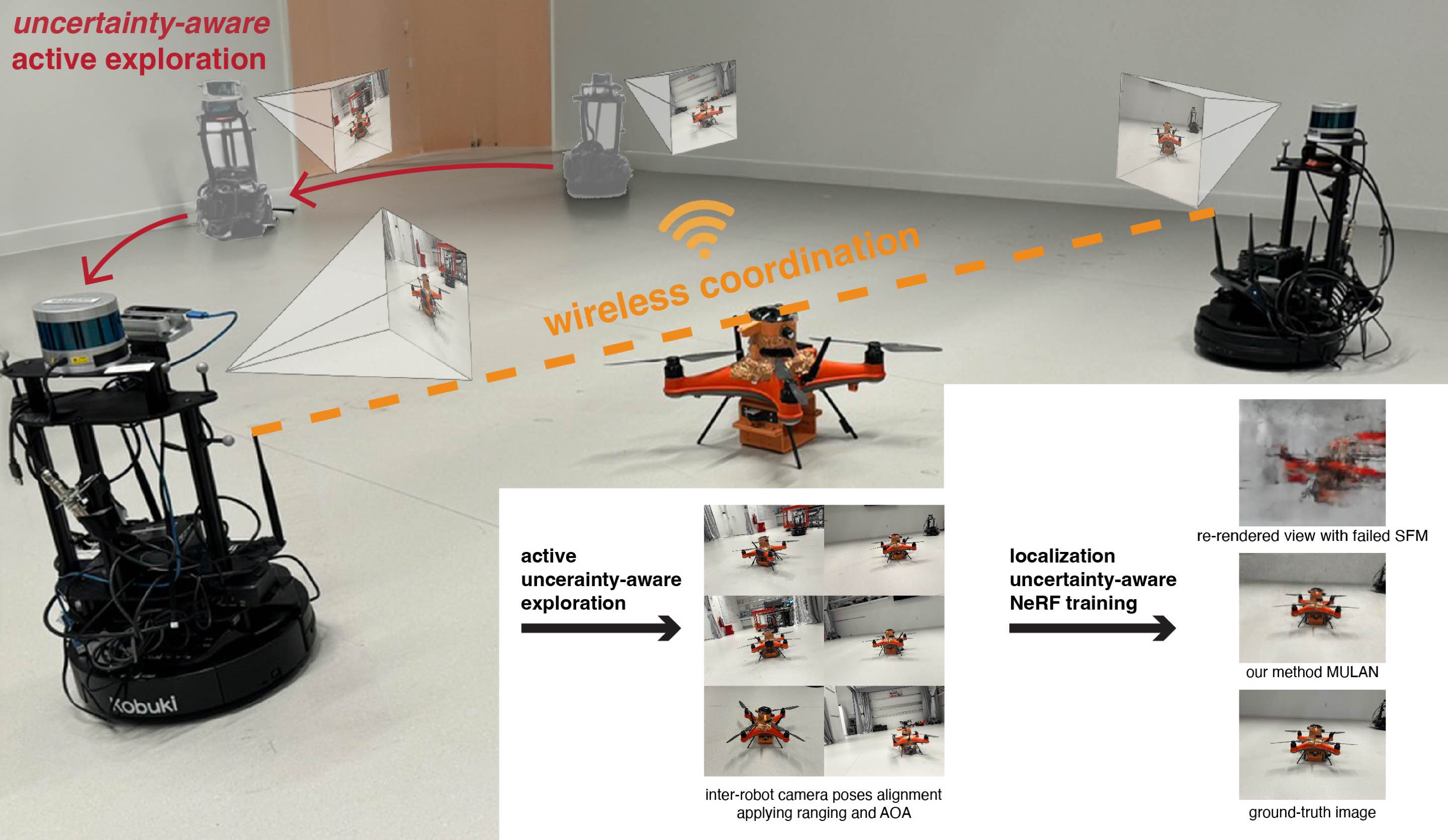}
           \captionof{figure}{Overview: We propose a collaborative, localization uncertainty-aware NeRF framework for a team of robots, employing wireless coordination and active best-next-view selection for novel view finding.}
           \label{fig:fig1}
        \end{center}
    }]
}

\author{\authorblockN{Weiying Wang\authorrefmark{1}, Victor Cai\authorrefmark{2}, Stephanie Gil\authorrefmark{1}}
\authorblockA{\authorrefmark{1}John A. Paulson School of Engineering and Applied Sciences\\
Harvard University \\ 
Email: weiyingwang, sgil@g.harvard.edu}
\authorblockA{\authorrefmark{2}victorcai@college.harvard.edu}
}


%

\maketitle

\begin{abstract}
This paper presents MULAN-WC, a novel multi-robot 3D reconstruction framework that leverages wireless signal-based coordination between robots and Neural Radiance Fields (NeRF). Our approach addresses key challenges in multi-robot 3D reconstruction, including inter-robot pose estimation, localization uncertainty quantification, and active best-next-view selection. We introduce a method for using wireless Angle-of-Arrival (AoA) and ranging measurements to estimate relative poses between robots, as well as quantifying and incorporating the uncertainty embedded in the wireless localization of these pose estimates into the NeRF training loss to mitigate the impact of inaccurate camera poses. Furthermore, we propose an active view selection approach that accounts for robot pose uncertainty when determining the best-next-
views to improve the 3D reconstruction, enabling faster convergence through intelligent view selection. Extensive experiments on both synthetic and real-world datasets demonstrate the effectiveness of our framework in theory and in practice. Leveraging wireless coordination and localization uncertainty-aware training, MULAN-WC can achieve high-quality 3D reconstruction that is close to applying the ground truth camera poses. Furthermore, the quantification of the information gain from a novel view enables consistent rendering quality improvement with incrementally captured images by commanding the robot to the novel view position. Our hardware experiments showcase the practicality of deploying MULAN-WC to real robotic systems. 
\end{abstract}
\section{Introduction}
Vision-based 3D reconstruction in previously unseen environments is pivotal in a broad spectrum of robotics applications, ranging from autonomous navigation \cite{meilland2015dense} to mapping and localization \cite{3dslam} to scene understanding \cite{nie2021rfd}. The conventional process typically involves: 1) collecting multi-modal sensory information from onboard sensors such as RGB-D cameras and inertial measurement units, 2) extracting geometric features to compute relative pose information, and 3) applying pose graph optimization to produce a 3D environment representation using geometrically constrained spatial feature information \cite{cao2018fast, chang2021kimera,yin2014graph}. Scaling up this capability to a fleet of robots could enable better coverage and faster exploration in large-scale environments. Nevertheless, it is nontrivial to scale up conventional methods to a fleet of robots. This is due to challenges in effectively obtaining relative poses between robots, which 
are needed to align inter-robot frames and form a global understanding of the scene. 
Another problem 
lies in how to actively command the robot to acquire visual information so as to maximize the information gain in the 3D reconstruction \cite{Gao2023mc-nerf,activRendezvous}.  Active image acquisition is even more critical in a multi-robot setting to fully leverage the advantages of the fleet over a single robot. To address these challenges, we introduce a multi-robot collaborative framework utilizing Neural Radiance Fields (NeRF) for reconstruction, and using on-board wireless signal-based coordination to provide relative positional information between robots.

Firstly, to address the need for photometrically and geometrically accurate 3D reconstruction, a large number of works in NeRF \cite{kerr2022evo,adamkiewicz2022vision,muller2022instant} offer a revolutionary technique in synthesizing photorealistic 3D implicit representations from sparse 2D images. This is attributed to NeRF's unique capability to model the volumetric density and color of light in a scene, enabling highly detailed and accurate reconstructions from diverse viewpoints. A crucial input that enables real-time NeRF in \cite{muller2022instant} is the camera pose corresponding to each image in the same frame. Using optical flow-based feature tracking described in \cite{rosinol2022nerfslam}, the relative camera poses can be computed in real-time and fed into the NeRF training. However, acquiring accurate relative positional information in a multi-robot system is nontrivial, especially in the absence of global localization systems like GNSS or motion capture systems. In the traditional multi-robot coordination or localization approach, multi-robot SLAM is often dependent on the alignment of individual maps and subsequent pose estimation from overlapped appearance-based feature observations \cite{chang2021kimera,do2020robust}, namely loop closure. However, inter-robot loop closure brings significant complications and computational overhead  \cite{wiclosure, chang2021kimera,active}. To satisfy the need for relative positional information, we instead use phased array-based wireless sensing between robots, building upon our previous work in \cite{wsr,toolbox}. Here, the off-the-shelf WiFi chip, which is native for communication on most robotics platforms, can be used to obtain inter-robot positional information independent of appearance-based environmental features. This positional information thus can be utilized to compute the relative translation between any pair of robots and thus their image frames. Inspired by other works in incorporating depth information from the SLAM pipeline \cite{rosinol2022nerfslam} where depth information is used as supervision, we also design our training loss to be aware of which regions of data are more certain than others based on the uncertainty in wireless sensing. Like other perception modalities, wireless sensing also encodes probabilistic perception due to environmental and hardware noise. This work develops a method to quantify the wireless localization uncertainty based on Angle of Arrival (AoA) profile reconstruction and correlation with the received AoA profile. 
Integrating this quantified uncertainty information into the NeRF training process allows us to bias the training loss and ensure that the training loss is informed about data regions with higher localization uncertainty, enhancing the accuracy and reliability of the 3D NeRF-generated reconstructions. 


A multi-robot system offers advantages beyond merely improving the efficiency of scene coverage from different viewpoints \cite{multi-robotslam}; we can also 
enable active image acquisition by determining the most informative next view for the NeRF model and controlling robots to acquire these additional images. Most works in applying NeRF to 3D reconstruction only passively process the images given by the perception pipeline. In resource-constrained large-scale deployment, it is beneficial to actively plan robots to acquire the most informative next image. 
Some works \cite{Pan2022ActiveNeRFLW,lee2023so} 
acquire images that can maximally cover the scene of interest, leading to higher information gain or better quality of reconstruction. However, to the best of our knowledge, this is the first time that active information acquisition has been applied to a coordinated multi-robot system for NeRF-based 3D reconstruction. The work in \cite{Pan2022ActiveNeRFLW} proposes a promising approach to evaluate the potential information gained from a novel view by quantifying the reduction of the variance for the rendering. However, this quantification does not consider the localization uncertainty of the camera, which is particularly 
necessary in multi-robot or multiple-camera setups, making it inefficient 
when dealing with localization uncertainty across robots that use wireless coordination. We address this specifically 
by considering the inter-robot camera pose uncertainty in the characterization of the color posterior in 3D space. Our work integrates localization uncertainty quantification into the evaluation of novel-view information gain by deriving the reduction of the variance. Subsequently, we can direct the robots to actively capture the best images from a set of feasible next positions, for the team of robots to achieve the highest information gain in the NeRF model.

In summary, this work makes three main contributions to multi-robot 3D reconstruction integrated with NeRF:
\begin{itemize}
    \item \textbf{Framework for integrating SAR-based wireless coordination for multi-robot NeRF
    }: We present a framework that leverages multi-robot collaboration and SAR-based wireless coordination to enable multi-robot localization uncertainty-aware NeRF. 
    \item \textbf{Collaborative active image acquisition}: Our system introduces a framework for active image acquisition, utilizing uncertainty quantification and novel-view location sampling to direct robots for optimal data collection, maximizing information gain for NeRF.
    \item \textbf{Extensive hardware experiments}: We conducted experiments on our customized hardware robot demonstrating that our framework does not only effectively achieve the same quality of rendering faster and higher converged quality, 
    but also can actively command the robot to an unvisited place that reduces the variance of rendering.
\end{itemize}

\section{Problem formulation}
In this section, we briefly review some background knowledge of NeRF and introduce the wireless coordination from our previous work \cite{toolbox} as a basis for our approach. 
\subsection{NeRF Formulation}
NeRF implicitly represents a scene using a fully connected neural network. In the ideal propagation ray-tracing model, the scene is modeled as a continuous function that maps any viewing angle $D$ of 5D input coordinates, consisting of the position in Cartesian coordinates $(x, y, z)$ and the viewing angle $(\theta, \phi)$, to a color $c (r,g,b)$ and a volume density $\sigma$. NeRF renders the color of the sampling ray passing through the environment with classical volume rendering. Suppose we sample a ray from a position $\boldsymbol{o}$ in direction $\boldsymbol{d}$. The points along the ray can be parameterized by 
\begin{align*}
    r(t) = \boldsymbol{o} + t\boldsymbol{d}
\end{align*}
The color projection of the ray back to the projection plane is 
\begin{align*}
    \mathcal{C}(r) = \int_{t_n}^{t_f}T(t)\sigma(r(t))c(D)\:dt,
\end{align*} 
where $T(t) = exp(-\int_{t_n}^t\sigma r(s)ds)$ is the accumulated transmittance along the sampling ray, and $t_n$ and $t_f$ are the artificial sampling box. In a realistic setup, the computation of the full integral of the color through the ray can be intractable. Instead, \cite{mildenhall2020nerf} discretizes the integral as the linear combination of multiple sample points along the ray. 
NeRF optimizes the approximated discrete function by minimizing the squared reconstruction error between the ground truth color of each pixel captured in training RGB images and the reconstructed rendering pixel colors. The loss function is then defined as 
\begin{align}
    \mathcal{L} = \sum_i||\mathcal{C}(r_i) - \bar{\mathcal{C}}(r_i)||^2_2
    \label{eq:nerf_loss}
\end{align}
where $\bar{\mathcal{C}}(r_i)$ is the captured color from images.
\subsection{Collaborative NeRF}
\label{sec:Collaborativenerf}
To achieve 3D reconstruction with more than one robot, one of the fundamental requirements is having a common frame of reference 
from a known camera extrinsic or relative transformation between cameras even if the data is collected from different robots from different views. Instead of being given a set of poses $\mathcal{T}$ in the same frame of reference, we instead focus on the problem of having the sets of poses from all the robots $\alpha, \beta,\dots$ in the team in $\mathcal{T}_\alpha$, $\mathcal{T}_\beta,\dots$. Without loss of generality, we only focus on the observation from two robots $\alpha$ and $\beta$. In order to align a pose $T_k^\alpha$ of robot $\alpha$ at local time k and another pose $T_p^\beta$ of robot $\beta$ at local time p, we need to obtain the inter-robot camera extrinsic $T_{k_\alpha}^{p\beta} = (t_{k_\alpha}^{p\beta}, \theta_{k_\alpha}^{p\beta})$, which is the distance and Angle of Arrival (AoA) between two cameras on different robots. 
\subsection{Wireless Coordination}
In our previous work \cite{wiclosure}, we extract the AoA information between any two robots by measuring the phase difference in the Wi-Fi channel. Suppose we have two robots $\alpha$ and $\beta$ in communicating range at time $t$ and their poses $T_\alpha$ and $T_\beta$ in local frames. We can measure relative position between two robots using ranging from the ultra-wideband (UWB) as well as AoA from our SAR-based framework output \cite{toolbox} with a probability density function of ranging and AoA annotated by $f_{uwb}(d|T^\alpha, T^\beta)$ and $f_{aoa}(\phi|T^\alpha, T^\beta)$ respectively, 
defined as:  
\begin{align}
    f_{uwb}(d|T^\alpha, T^\beta)&= c_{1} \exp{\left(\sigma_{k,p}^{-2} (d-\|t_{k\alpha}^{p\beta} \|_2)^2\right)} \label{eq:factor_ranging}\\
    f_{aoa}(\phi|T^\alpha, T^\beta)&= c_{2}\exp{\left(\kappa_{k,p}^2cos(\phi_{k\alpha}^{p\beta} - \theta_{k\alpha}^{p\beta})\right)} \label{eq:factor_aoa}
\end{align}
where $c_{1}=\frac{1}{\sqrt{2\pi\sigma_{k,p}^2}}$ and $c_{2}=\frac{1}{2\pi I_0(\kappa_{k,p})}$. 
Here, $\sigma_{k,p}^2$ and $t_{k\alpha}^{p\beta}$ are the variance and mean of the distance measurement; and $\kappa^2_{k,p}$ and $\theta_{k\alpha}^{p\beta}$ are the concentration parameters computed as the inverse of the AoA variance and the mean of the AoA distribution.

\section{Approach}
In this section, we present our multi-robot NeRF framework that addresses the challenges of inter-robot pose localization, uncertainty quantification, and active best-next-view finding. Our approach leverages wireless signals, specifically ranging and Angle of Arrival (AoA) measurements, to estimate the relative poses between robots. We develop a novel method to quantify the uncertainty of AoA estimates by reconstructing the AoA profile and correlating it with the received AoA profile. This uncertainty quantification is then integrated into the NeRF training process to mitigate the impact of inaccurate poses on the reconstruction quality. Furthermore, we propose an active view-finding approach that accounts for the position uncertainty of the robots when selecting the most informative views for NeRF training. By incorporating localization uncertainty into the novel view selection process, our framework can more accurately determine the best next views for each robot, even in the presence of pose uncertainty arising from wireless coordination.
\subsection{Inter-robot Pose Localization and Uncertainty Quantification}\label{equ}
As described in Section~\ref{sec:Collaborativenerf}, 
accurate multi-robot NeRF reconstruction relies 
on 
obtaining the transformation between cameras on different robots. In a multi-robot setup, we propose using wireless signals leveraging UWB ranging and WiFi AoA measurements to obtain accurate inter-robot poses. Suppose we have a pose $T_k^\alpha$ in SE(3) of robot $\alpha$ at local time k and another pose $T_p^\beta$ of robot $\beta$ at local time p. We then can obtain a wireless measurement of range and AoA between the two robots using onboard UWB and WiFi 
annotated by a tuple $(t_{k_\alpha}^{p\beta}, \theta_{k_\alpha}^{p\beta})$. If we aim to use robot $\alpha$'s frame as the global frame, then the extrinsic or the rigid transformation of robot $\beta$'s camera pose can be represented by $ T_{kp}^{\alpha\beta} \oplus  T_p^\beta$,
 where the annotation $\oplus$ denotes rigid transformation. However, the accuracy of the resulting pose estimate is subject to the uncertainties in the ranging and AoA measurements as described in Eq~\ref{eq:factor_ranging} and Eq~\ref{eq:factor_aoa}. To mitigate the impact of inaccurate poses on the NeRF training process, we propose applying a weight to each training example based on the uncertainty of the associated robot pose's ranging and AoA measurements from the other robots.


Quantifying the uncertainty of AoA estimates is particularly challenging, since there is a lack of standard error quantification methods applicable from previous works. To address this, we propose a novel approach 
as follows.
The ideal channel on wavelength $\lambda$ at robot $\alpha$ from robot $\beta$ over distance $d_{\alpha\beta}(t)$ is 
\begin{align}h_{\alpha\beta}(t) = \frac{1}{d_{\alpha\beta}(t)} exp\left(\frac{-2\pi 
\sqrt{-1}}{\lambda} d_{\alpha \beta}(t)\right)\end{align}

Suppose over $t=t_k,\dots,t_l$, robot $\alpha$ receives the measured channel $\overline{h}_{\alpha\beta}(t)$ and also collects its local pose information $\overline{T}_{\alpha}(t)$ 
containing the displacement distance, azimuth, and zenith of robot $\alpha$ from the center of its frame. Then the measured AoA profile is constructed as in \cite{wsr} to be
$\overline{F}_{\alpha\beta}(\phi,\theta)$ over a sample space in $(\phi,\theta)$
and the measured AoA is chosen as $(\overline{\phi},\overline{\theta}) = \arg \max_{(\phi,\theta)}\left\{\overline{F}_{\alpha\beta}(\phi,\theta)\right\}$ at the tallest peak.

\begin{figure}[h]
    \centering
    \vspace{-0.3cm}
\includegraphics[width=0.5\textwidth]{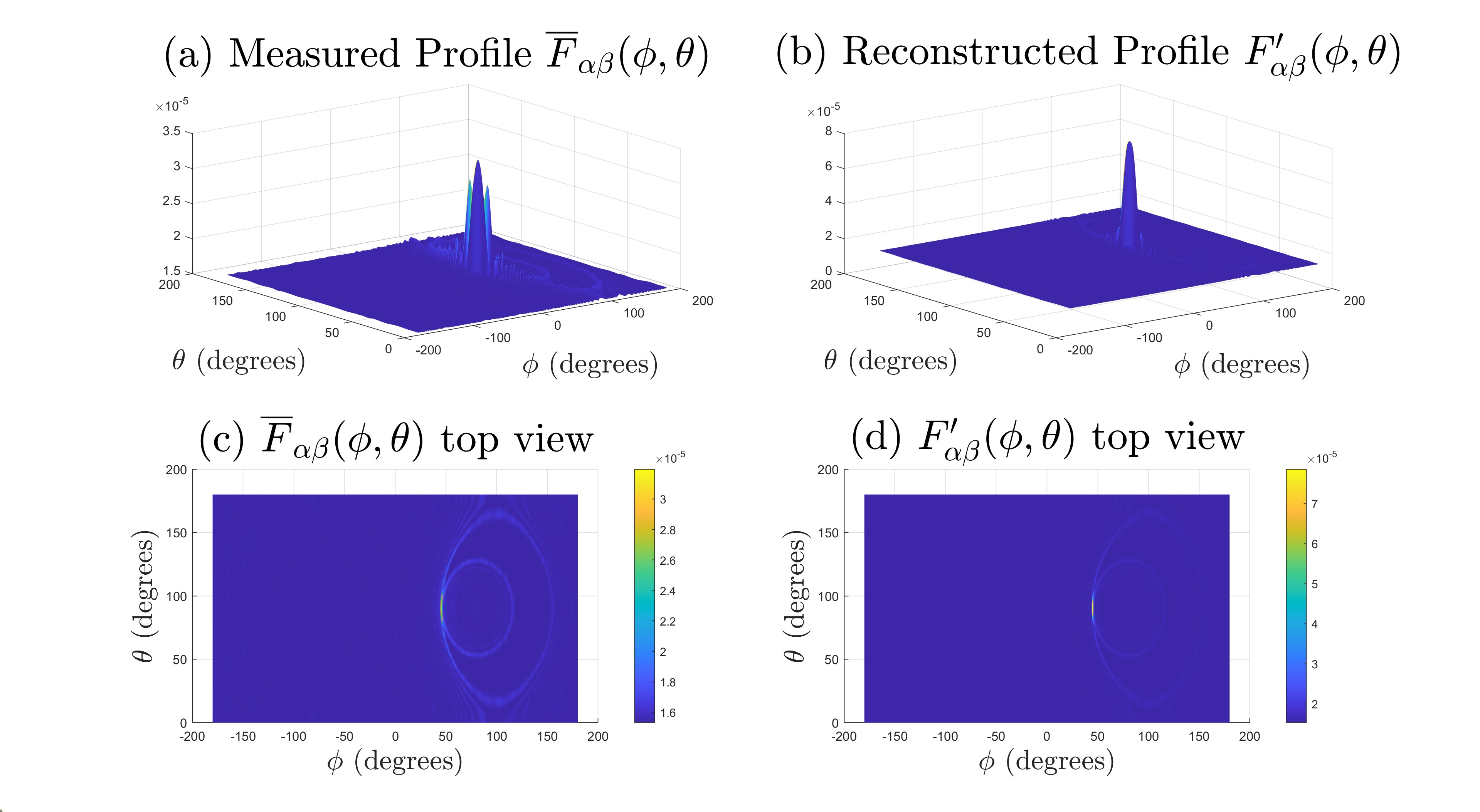}
    \caption{Simulation of the AoA variance methodology. (a) Measured AoA profile $\overline{F}_{\alpha\beta}(\phi,\theta)$ from a simulated measured wireless channel $\overline{h}_{\alpha\beta}$ with $0.7$ radians standard deviation injected phase noise. (b) Reconstructed profile $F_{\alpha\beta}'(\phi,\theta)$ from reconstructed channel $h_{\alpha\beta}'$, with $0.5$ radians standard deviation phase noise for tolerance. Both tallest peaks align at $(\overline\phi = 45.6^{\circ},\overline\theta=90^{\circ})$. (c) and (d) are respective top views.}
    \label{fig:compare_measured_and_reconstructed_profiles}
    \vspace{-0.1cm}
\end{figure}

Now, given $(\overline\phi,\overline\theta)$ and pose information $\overline{T}_\alpha(t_k),\dots,\overline{T}_\alpha(t_l)$, we reconstruct the channel over $t=t_k,\dots,t_l$ as
\begin{align}\label{channel noise} h_{\alpha\beta}'(t) = exp\left(\frac{-2\pi 
\sqrt{-1}}{\lambda} f_\alpha(\overline{T}_\alpha(t),\overline\phi,\overline\theta)+\nu(t)\sqrt{-1}\right)\end{align}
where $f_\alpha$ is the displacement of robot $\alpha$ projected along the measured AoA direction in the local frame, relative to the first observation at $t_k$, and
$\nu(t)$ is a zero-mean real random variable that injects Gaussian phase noise into each element of the channel to add a small realistic amount of error tolerance in the measured profile. The same AoA algorithm is run to obtain the reconstructed profile $F_{\alpha\beta}'(\phi,\theta)$ and its tallest peak $(\phi',\theta')$. Since $F_{\alpha\beta}'(\phi,\theta)$ is constructed from $(\overline\phi,\overline\theta)$, making the noise $\nu(t)$ small ensures that the reconstruction has $(\phi',\theta')=(\overline\phi,\overline\theta)$ when the $\phi$ and $\theta$ sample spaces are discretized during profile computation. Figure  \ref{fig:compare_measured_and_reconstructed_profiles} simulates the profiles for illustration.

The AoA uncertainty is then calculated as follows. Both profiles are cropped to a small rectangle $R$ with $\Delta\phi=\Delta\theta=10^\circ$ around the measured AoA $(\overline\phi,\overline\theta)$ to ignore distant multipath components. Then define AoA uncertainty $\kappa_{k,p}$ via
\begin{align} \frac{1}{\kappa_{k,p}} =  \sum_{(\phi,\theta)\in R} \overline{F}_{\alpha\beta}(\phi,\theta) F_{\alpha\beta}'(\phi,\theta) \end{align}

This measures how concentrated the received profile is around its peak $(\overline\phi,\overline\theta)$, such that a lower AoA uncertainty $\kappa_{k,p}$ corresponds to a more reliable AoA measurement. This AoA variance can be used in Equation \ref{eq:factor_aoa} to understand the variability in the pose estimates, as described in the next section.

\subsection{Uncertainty-aware NeRF Training}
\label{sec:Uncertainty nerf}
As described in our problem formulation, if we have two sets of camera poses $T_\alpha$ and $T_\beta$ in their own local frames, we want to find the relative transformation between them to align the poses in the same frame of reference. For each wireless measurement between 
poses $T_k^\alpha$ and $T_p^\beta$, we can quantify the uncertainties $\sigma_{k,p}$ and $\kappa_{k,p}$ respectively of the ranging and AoA estimates using the methods described in the previous section.
We propose incorporating these uncertainty measures into the NeRF training process by modifying the standard pixel loss function $\mathcal{L}$ given in Eq~\ref{eq:nerf_loss} by re-scaling the loss for each training sample. For brevity of the annotation in this section, we omit robot indices $\alpha$, $\beta$, and all local time frames p and k from now on. We apply uncertainty propagation to compute the error ellipse of uncertain measurements. Since the ranging and AoA measurements are taken using different sensing modalities, they can be treated as independent measurements with the error ellipse's axes aligning with x-y axes. The semi-major and semi-minor axes $a$ and $b$ of the error ellipse can be derived as:
\begin{align}
    a^2 = \sigma^2 cos^2(\theta) + t^2 sin^2(\theta)\kappa^2 \\
    b^2 = \sigma^2 sin^2(\theta) + t^2 cos^2(\theta)\kappa^2
\end{align}
where $t$ is the ranging measurement and $\theta$ is the AoA estimate.
Then the scale factor with confidence interval $CI$ is given by 
\begin{align}
    k = \sqrt{-2log(1-CI)}
\end{align}

Hence, the uncertainty of this wireless localization $\gamma$ can be represented by the area of the error ellipse $\gamma = k^2\pi ab$. Then the new loss function can be re-scaled with $\gamma$ that is normalized by sigmoid function
\begin{align}
    \mathcal{L}_{uncertainty} = \textit{SIGMOID}(\gamma)\mathcal{L}
    \label{eq:adaptive_loss}
\end{align}

By incorporating the uncertainty-aware scaling factor into the NeRF loss function, our multi-robot NeRF system can effectively learn to reconstruct the 3D scene while accounting for the varying reliability of the pose estimates obtained through wireless coordination. This approach results in a more robust and accurate 3D reconstruction, especially in scenarios where the pose estimates may be subject to significant uncertainties.
\vspace{-0.6cm}
\subsection{Active Best-View Finding with Position Uncertainty}
In a multi-robot NeRF system, actively selecting the best views for each robot to capture can significantly improve the efficiency and quality of the 3D reconstruction. However, the uncertainty in robot poses obtained through wireless coordination can impact the effectiveness of the view selection process. When a robot attempts to find the best next view location by proposing and evaluating potential new positions, the uncertainty in its current pose can lead to inaccurate assessments of the information gain at 
novel view locations.

To address this challenge, we propose an active view finding approach that incorporates the positional uncertainty of the robots to guide the selection of the most informative views for NeRF training. Building upon the approach proposed in \cite{Pan2022ActiveNeRFLW} for evaluating the potential information gain from novel views by quantifying the reduction of variance in rendering, we extend this method to account for the uncertainty in the robot's current position and its propagation to the novel view locations being evaluated. By considering localization uncertainty during the novel view selection process, we can more accurately determine the most informative next views for each robot, even in the presence of pose uncertainty arising from wireless coordination.

We adopt the assumption that the radiance color of any location along the ray $r(t)$ can be parameterized by a Gaussian distribution with mean $\bar{c}(r(t))$ and variance $\bar{\beta}(r(t))$.
To incorporate this uncertainty, we model the origin $\boldsymbol{o}$ of each ray $r$ following a Gaussian distribution with $0$ mean and the variance $\sigma$, representing the localization uncertainty.
\begin{align}
    \boldsymbol{o} \sim \mathcal{N}(0,\sigma)
\end{align}
Assuming a NeRF model $\mathcal{M}$ has been trained on an initial collection of data $D$, the prior distribution $P(c(r(t_k))|D)$ of the color $c$ at location $r(t)$ follows a Gaussian distribution $\mathcal{N}\sim (\bar{c}(r(t)), \bar{\beta}^2(r(t)))$.
The accumulated color from a new ray $r$ passing through can also be modeled as a Gaussian distribution: 
\begin{align}
p(C(r)|c(r(t)), \boldsymbol{o}) 
\end{align}
where $C(r)$ is the color of the rendered pixel accumulated from the ray $r$, and $c(r(t))$ is the color of the location in 3D space.
Then if we marginalize over $\boldsymbol{o}$, 
\begin{align}
     p(C(r)|c(r(t))) &= \int p(C(r)|c(r(t_k))) * p(\boldsymbol{o}) \: d\boldsymbol{o}
\\
p(C(r)|c(r(t_k))) &
\sim \mathcal{N}(\sum_{i=1}^N\alpha_i\bar{c}(r(t)),\sum_{i=1}^N\alpha_i*\sigma^2+\bar{\beta}^2(r(t_k))
)
\end{align}
Then apply Bayes' rule to get the posterior:
\begin{align}
    &P(C(r)|D,r(t_k),\boldsymbol{o} ) \notag \\
   & \propto P(C(r)|c(r(t_k))) * P(c(r(t_k))|D) * P(\boldsymbol{o}) \notag \\ 
   & \propto exp\left(
   \vphantom{\left(\frac{\alpha_k^2}{\alpha_k^2\sigma^2 + \bar{\beta}^2(r)}+\frac{1}{\bar{\beta}^2(r(t_k))}\right)^{-1}} 
   -\frac{1}{2} \left(c(r(t_k)) -  \left(\frac{\omega  C(r) }{ \alpha_k }+ (1 - \omega)\right) * \bar{c}(r(t_k))\right) * \notag  \right. \\ &\left. \left(\frac{\alpha_k^2}{\alpha_k^2\sigma^2 + \bar{\beta}^2(r)}+\frac{1}{\bar{\beta}^2(r(t_k))}\right)^{-1}\right) \notag 
 \\
   & \textit{where } \omega = \frac{\alpha_k^2\bar{\beta}^2(r(t_k))}{\alpha_k^2\bar{\beta}(r(t_k))^2+\alpha_k^2\sigma^2+\bar{\beta}^2(r)}
\end{align}
Then extract the variance of the posterior distribution:
\begin{align}
\left(\frac{\alpha_k^2}{\alpha_k^2\sigma^2 + \bar{\beta}^2(r)}+\frac{1}{\bar{\beta}^2(r(t_k))}\right)^{-1}
\label{eq:variance}
\end{align}

As the localization variance increases, the uncertainty of the radiance field also increases accordingly. To select the best view, the metric will prefer the novel view with lower localization uncertainty. Since we only need to consider the variance reduction given multiple rays from a sampled novel-view position, we can then command the robot to move to the location with highest variance reduction using Equation~\ref{eq:variance}. 

\section{Results}
In this section, we present a comprehensive evaluation of our methodology through both synthetic datasets collected in synthetic environments and real-world datasets collected on our hardware robots. Our findings validate the effectiveness of our algorithm in integrating perspectives from multiple robots within an active acquisition framework, showcasing significant improvements in data capture and processing.
We implemented our algorithm based on a Pytorch implementation of the SDF and NeRF part described in Instant-NGP \cite{muller2022instant} with CUDA-accelerated ray marching. We modified the loss function to implement the localization uncertainty-aware loss described in Eq~\ref{eq:nerf_loss}. Further, for the active image collection, we re-implemented the rendering variance reduction described in \cite{Pan2022ActiveNeRFLW}, incorporating the localization uncertainty from Eq~\ref{eq:variance}. We use a desktop with NVIDIA RTX 6000 for all evaluations.
\subsection{Wireless Variance}
\begin{figure}[h]
       \vspace{-0.2cm}
    \centering
\includegraphics[width=0.45\textwidth]
{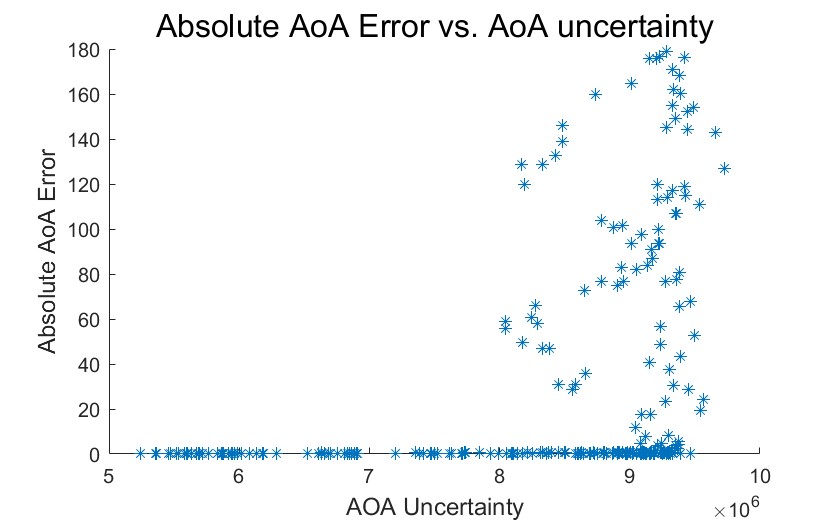}
    \caption{Absolute AoA error from ground truth plotted against our AoA uncertainty metric. We see that nonzero AoA error grows as our AoA uncertainty metric grows, indicating that our metric successfully captures true error in measured AoA.}
    \label{fig:variance_vs_error}
\end{figure}

\begin{figure}[h]
    \centering
    \vspace{-0.35cm}
\includegraphics[width=0.45\textwidth]{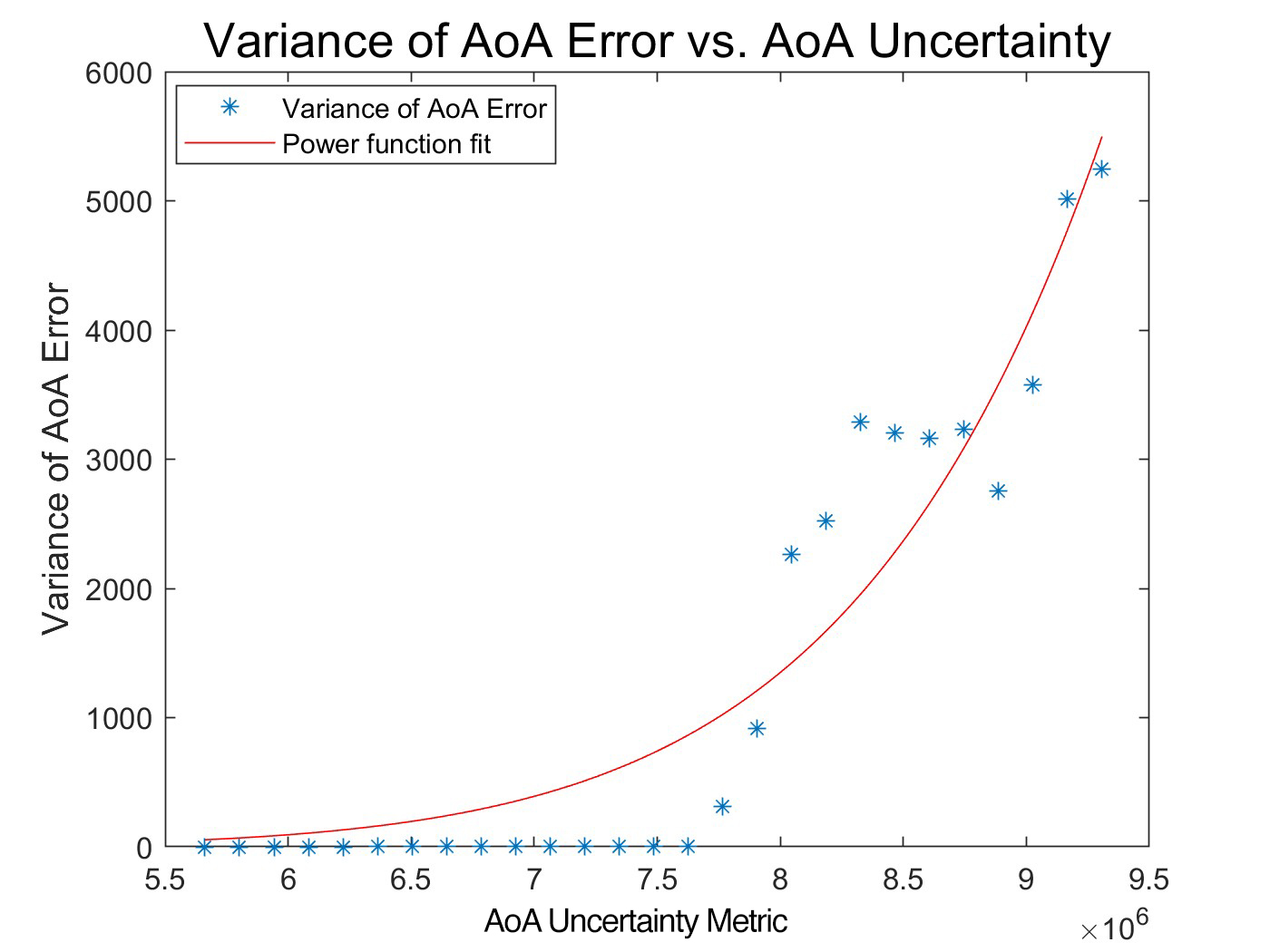}
    \caption{The variance of the AoA error here as a function of AoA uncertainty is calculated empirically by finding the variance of the AoA error on the y-axis within a sliding window of $\Delta\kappa_{k,p} = 8.4\times 10^5$ along the x-axis. It is fit with a power curve of the form $y=ax^b$, with $r^2=0.8942$.}
    \label{fig:channnel-variance}
           \vspace{-0.6cm}
\end{figure}

First, we present the performance benchmark with our proposed wireless variance metric as formulated in Section \ref{equ}. With a simulated trajectory, the metric is tested over $16$ random trials with various amounts of injected Gaussian channel phase noise as defined in Equation \ref{channel noise}, ranging from $0.01$ to $3$ radians standard deviation. As shown in Figure~\ref{fig:variance_vs_error}, the AoA error from the ground truth scales quickly and becomes unstable as our proposed AoA variance metric increases. To further demonstrate the relationship between the variance of the AoA error and our proposed profile variance, Figure~\ref{fig:channnel-variance} clearly shows that the variance of the AoA error will increase superlinearly as our metric increases. This result indicates that our AoA uncertainty quantification is well aligned as an indicator of the variance of AoA measurements, which can be further used to quantify the uncertainty of the camera position. 

\subsection{Simulation Experiment}
Our 
localization uncertainty-aware framework described in Sec~\ref{sec:Uncertainty nerf} is first assessed using a synthetic dataset lego released with the original NeRF work \cite{mildenhall2020nerf}
 that is 
commonly used for evaluating NeRF frameworks. The dataset is partitioned into two subsets to simulate data acquired from two robots, allowing 
us to mimic the real-world scenario of capturing images from different angles and positions, thereby testing the robustness and adaptability of our algorithm in synthesizing and analyzing data from varied viewpoints. Three setups are evaluated with results in Table~\ref{table:1}:
\begin{enumerate}
\item [A] [Oracle]: Camera poses from the dataset in
the global frame (known as extrinsic between cameras) and
images from the dataset.
\item [B] [Normalized camera poses with AOA and ranging simulated]: Camera poses from the dataset but normalized by the first pose in each partition. Then the AOA and ranging are simulated with noise whose standard deviation are 0.05 meters and 5 degrees respectively. 
 \item [C] \textbf{[Normalized camera poses with AOA and ranging simulated with variance as supervision]:} The same setup as B but the training loss is incorporated with the localization 
 variance. 
\end{enumerate}
As predicted, training using localization variance produces better performance closer to using ground truth poses.

\begin{table}[h]
\centering
\begin{tabular}{||c c c||} 
 \hline
  \multirow{2}{2em}{A}  & PSNR &  30.47  \\
   & LPIPS &0.062  \\
   \hline
 \multirow{2}{2em}{B}  & PSNR &  26.48 \\
   & LPIPS & 0.092  \\
 \hline 
 \multirow{2}{2em}{C}  & PSNR &  28.69  \\
   & LPIPS & 0.071   \\
   \hline
\end{tabular}
\caption{Performance comparison between different setups where larger PSNR values are better, and smaller LPIPS values indicate better quality
, illustrating that applying uncertainty-aware loss can effectively improve the quality of the model.}
\label{table:1}
\vspace{-0.3cm}
\end{table}

\subsection{Hardware Experiment}
For a real-world application, we deployed our algorithm on two customized Locobot PX100 robots. These robots were equipped with Oak-D Pro cameras, operating at 1080p 20Hz, along with DWM1001 UWB modules, 5dBi Antennas, and Intel NUC 10 computers for onboard processing. The experimental setup places a drone as a test object central relative to the two robots, which are programmed to navigate curved paths around the object to complete data capture. The wireless AOA measurements are computed by deploying \cite{toolbox} which only requires very small communication bandwidth at around 5 kB/s.

Both robots utilize onboard Visual Inertial Odometry (VIO) to estimate local camera displacement within their respective frames. At the onset of the experiment, Angle of Arrival (AoA) and ranging measurements are taken to establish an initial estimate of the relative positioning between the robots. Subsequently, the covariance of the VIO data was monitored to identify optimal intervals for refreshing wireless data collection. In the meantime, the testbed is equipped with the Optitrack motion capture (mocap) system providing the ground truth camera poses for each robot. 

The experiments are conducted using five setups:
\begin{enumerate}
    \item[A] \textbf{[Oracle]:} Camera poses captured by motion system in the global frame (known extrinsic between cameras) and images from the onboard camera.
    \item[B] \textbf{[Best case for our system]:} Camera poses from motion capture system with wireless coordination and images from the onboard camera. The poses are normalized in each robot's local frame.
    \item[C] \textbf{[Our system ``in-the-wild'' (no mocap)]:} Camera poses from onboard VIO, wireless coordination, and images from the onboard camera. 
    \item[D] \textbf{[Our system ``in-the-wild'' with variance as supervision]:} Camera poses from onboard VIO, wireless perception for coordination, and uncertainty-aware training loss scaled by localization variance.
    \item[E] \textbf{[Benchmark comparison]:} Camera poses from on-board VIO; COLMAP \cite{colmap} is used for computing inter-robot relative camera pose extraction.
\end{enumerate}
All five setups are evaluated using standard metrics for NeRF: Peak Signal-to-Noise Ratio (PSNR) and Learned Perceptual Image Patch Similarity (LPIPS) \cite{lpips}. Each metric is evaluated from samples in the test set ground truth images, along with camera poses. For each setup, there are a total of 100 images with camera poses that are collected continuously from each robot while robots are moving around the drone subject. drone-1 an drone-2 are different images in the testing dataset.
\begin{table}[h]
\centering
\begin{tabular}{||c c c c ||} 
 \hline
  & & drone-1 & drone-2 \\
 \hline
  \multirow{2}{2em}{A}  & PSNR &  26.4 & 24.5 \\
   & LPIPS &0.351  & 0.384  \\
   \hline
 \multirow{2}{2em}{B}  & PSNR &  25.45 & 23.4  \\
   & LPIPS &0.382 & 0.378   \\
 \hline 
     \multirow{2}{2em}{\textbf{C}}  & \textbf{PSNR} &  \textbf{23.32} & \textbf{22.3}   \\
   & \textbf{LPIPS} &\textbf{0.41} & \textbf{0.405}  \\
\hline
\multirow{2}{2em}{\textbf{D}}  & \textbf{PSNR} &  \textbf{25.04} & \textbf{23.03} \\
   & \textbf{LPIPS} & \textbf{0.389} & \textbf{0.395} \\
\hline
\multirow{2}{2em}{E}  & PSNR &  11.5 & 12.5  \\
   & LPIPS &0.79 & 0.85 \\
 \hline
\end{tabular}
\caption{Performance comparison between different setups where larger PSNR values are better, and smaller LPIPS values indicate better quality. The comparison between setup A and setup B demonstrates that applying wireless coordination can effectively achieve close performance to having a global coordinate system. Results from setup C show the realistic performance of our system using fully onboard VIO for local positioning, which is degraded but still relatively robust. Setup D shows the scaling with localization uncertainty quantification can improve the quality almost to the best-case scenario in setup B. Setup E fails to produce a coherent 3D rendering.}
\label{table:full_res}
\vspace{-0.1cm}
\end{table}

As shown in Table~\ref{table:full_res} which provides our quantitative results, setup A shows the best performance we can achieve in a two-robot team since it is based on the ground truth camera poses provided by the motion capture system.
In setup B, the poses are normalized by the starting pose of each robot captured in the motion capture system. Then wireless coordination is incorporated to provide inter-robot camera extrinsic. 
Setup C shows the realistic setup, which applies the wireless localization between robots to the local VIO poses and is effectively close to the result in setup B. 
Setup D shows that our framework can achieve better results than C by using the variance-aware loss function defined in Equation \ref{eq:adaptive_loss} with the corresponding localization variance proposed in Equation \ref{eq:variance}. 
The benchmark comparison setup E fails to produce a cohesive 3D rendering due to discrepancies in the relative camera pose estimation using COLMAP \cite{colmap}, which is commonly used for estimating the relative camera pose given two frames from different views. This is mainly due to the drastic translation change in camera view from different robots. This results also suggests that COLMAP won't be a proper solution for a multi-robot setup.
\begin{figure}
    \centering
   \includegraphics[width=.5\textwidth]{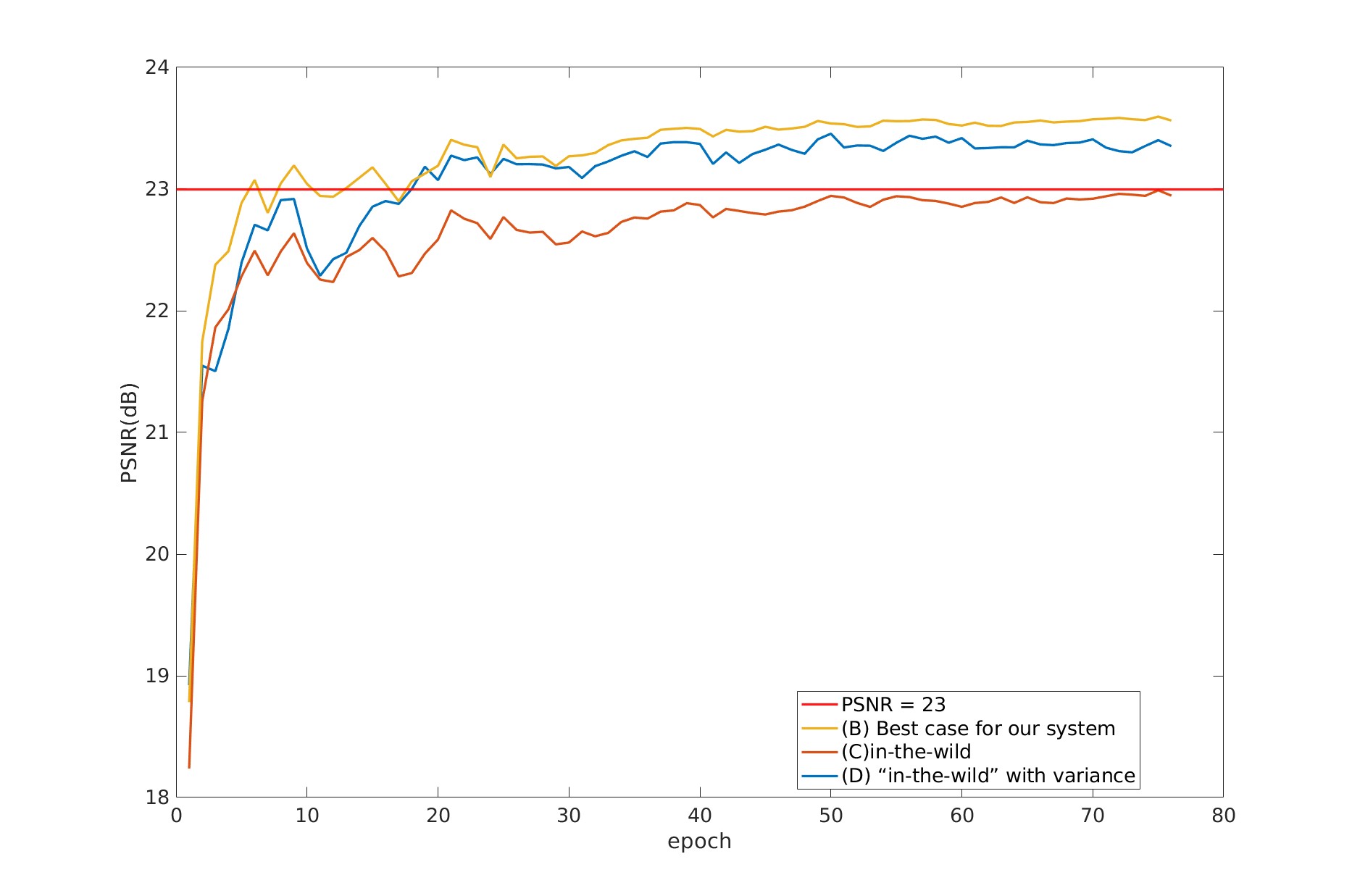}
       \vspace{-0.75cm}
    \caption{PSNR improvement over epochs, 
    with different setups.}
    \label{fig:PSNR}

\end{figure}

Many robotics applications require a quickly-converging view of the environment before the model training fully converges. In our experiment, we also validate that our methods not only deliver better rendering but also achieve faster PSNR improvement as shown in Fig~\ref{fig:PSNR}.

\begin{figure}
       \vspace{-0.15cm}
     \begin{subfigure}{0.24\textwidth} 
    \includegraphics[width=\textwidth]{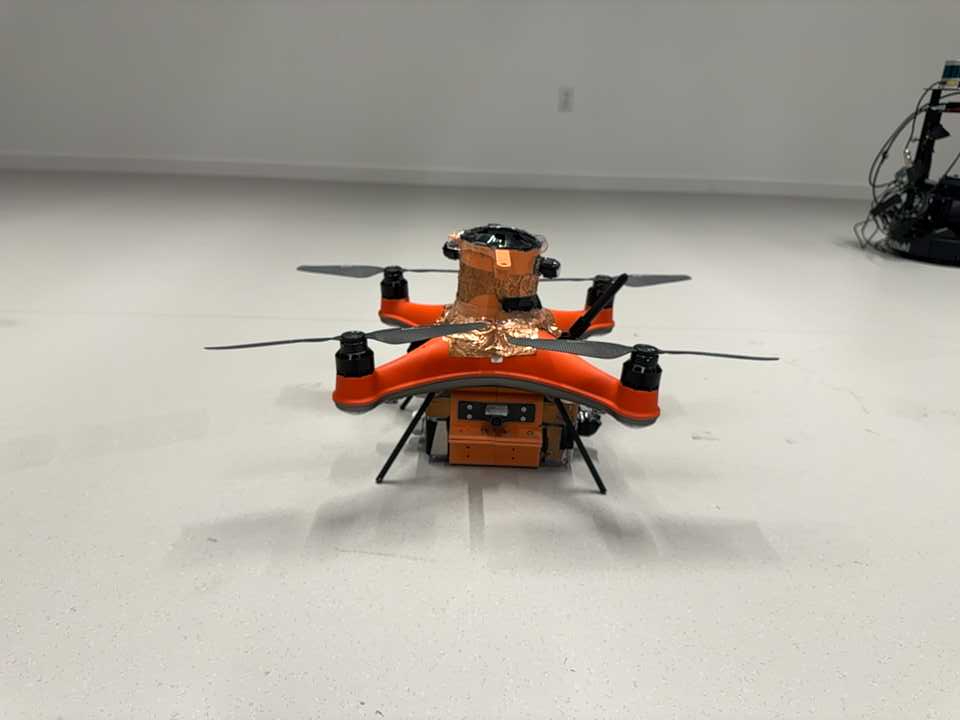}
    \caption{Ground truth drone image.\\ \; \\ \; 
    }
         \end{subfigure}
     \begin{subfigure}{0.24\textwidth} 
    \includegraphics[width=\textwidth]{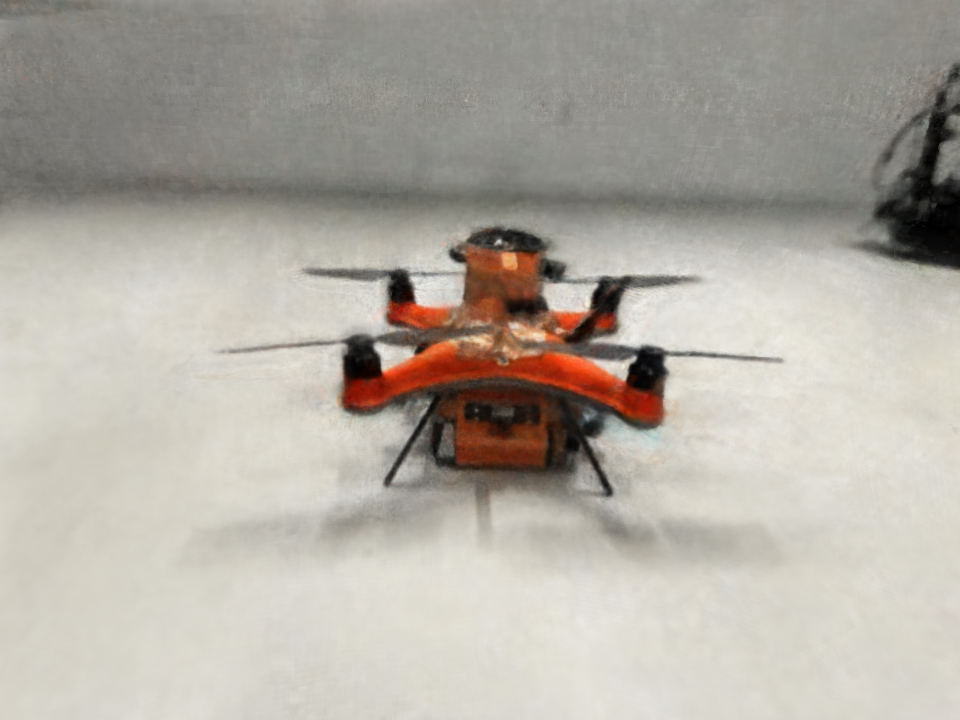}
    \caption{Re-rendered drone using wireless coordination and uncertainty-aware loss(PSNR:25.04).}
         \end{subfigure}
    \caption{An example of the drone we reconstructed in the testbed. The left figure is the ground truth image, right figure is the re-rendered image from a trained model.}
           \vspace{-0.25cm}
\end{figure}
\subsection{Active Image Capturing}
For evaluating active best next view, following the initial phase of data gathering, a waiting period was observed until the Neural Radiance Field (NeRF) model's loss stabilized. The robots then execute a series of maneuvers, sampling eight different directions at 0.5-meter intervals. Our evaluation focused on minimizing the variance of the rendering posterior, employing Equation \ref{eq:variance} to identify positions yielding the most significant reduction in variance.
The application of our algorithm in a hardware setting demonstrates its practical feasibility. Moreover, it underscores the potential of our method to optimize the data capture process through strategic robot positioning and movement.

After selecting the location with the highest variance reduction using our proposed method, the robot is commanded to the new location and observes the environment again. We then let the model train until the loss stabilizes and repeat the process four times to evaluate the efficacy of our method. For comparison, we also randomly selected accessible locations around the robots and controlled the robots to move to those locations. The evaluation-maneuver-training loop was conducted on both our policy and the random policy and the results are reported in Table~\ref{table:active_res}. These results demonstrate that our approach provides a principle metric that can improve the quality of the rendering consistently.

\begin{table}[h]
\centering
\begin{tabular}{||c c c c c c ||} 
 \hline
observation\# & & 1st & 2nd & 3rd & 4th  \\
 \hline
  \multirow{2}{4em}{Our algorithm}  & PSNR &  19.66 & 19.80 & 20.04 & 20.08  \\
   & LPIPS &0.407 & 0.398 & 0.394 & 0.381  \\
   \hline
 \multirow{2}{4em}{Random Exploration}  & PSNR &  19.53 & 19.63 & 19.60 & 19.63  \\
   & LPIPS &0.422 & 0.419 & 0.421 & 0.418  \\
 \hline
\end{tabular}
\caption{Performance comparison between different setups, demonstrating that our method improves the rendering quality metric with consecutive views.}
\label{table:active_res}
\vspace{-0.5cm}
\end{table}
\section{Conclusion} 
\label{sec:conclusion}
This work presents MULAN-WC, a multi-robot 3D reconstruction method that uses wireless signal-based coordination between robots. This work presents i) a framework for multi-robot NeRF that uses SAR-based wireless relative position measurements to stitch together views of the environment from multiple robots, ii) uncertainty-based weighting of samples in the NeRF training as a supervision technique, where samples with greater wireless measurement noise are weighted less, leading to better accuracy of the combined rendering, and iii) collaborative active next-image acquisition, where novel-view location sampling incorporates wireless pose uncertainty, and is used to direct robots to better sampling locations that reduce variance during NeRF training. We demonstrate the performance of the multi-robot framework in hardware, where our results show good quality of rendering according to the standard NeRF error metrics of PSNR and LPIPS, and consistent improvement when we additionally use the uncertainty of the AoA measurement as supervision in the NeRF training. Lastly, we show that AoA measurements can be used to select the best-next-view 
based on regions of better position accuracy and that this results in incremental rendering quality improvement.

\section*{Acknowledgments}
The authors gratefully acknowledge partial funding through NSF grant $\#$CNS-2114733, Amazon ARA, and the Sloan award $\#$FG-2020-13998.


\bibliographystyle{IEEEtran}
\bibliography{references}

\end{document}